\pdfoutput=1

\documentclass[11pt]{article}

\usepackage[]{ACL2023}

\usepackage{times}
\usepackage{latexsym}

\usepackage[T1]{fontenc}

\usepackage[utf8]{inputenc}

\usepackage{microtype}

\usepackage{graphicx}
\usepackage{subfigure}
\usepackage{amsmath} 
\usepackage{amsfonts}
\usepackage{algorithm}
\usepackage{algpseudocode}
\usepackage{multirow}
\usepackage{enumitem}

\title{Annotation-Inspired Implicit Discourse Relation Classification with Auxiliary Discourse Connective Generation}

\author{
Wei Liu \and Michael Strube \\ 
Heidelberg Institute for Theoretical Studies gGmbH  \\ 
\texttt{\{wei.liu, michael.strube\}@h-its.org}
}

\begin{document}
\maketitle

\begin{abstract}
Implicit discourse relation classification is a challenging task due to the absence of discourse\- connectives. To overcome this issue, we design an end-to-end neural model to explicitly\- gene\-rate discourse connectives for the task, inspired by the annotation process\- of PDTB. Specifically\-, our model jointly learns to gene\-rate discourse connectives between arguments\- and predict discourse relations based on the argument\-s and the generated connect\-ives. To prevent our relation classifier from being misled by poor connectives generated at the early stage of training while alleviating\- the discrepancy between training and inference, we adopt Schedule\-d Sampling to the joint learning. We evaluate our method on three benchmarks, PDTB 2.0, PDTB 3.0, and PCC. Results show that our joint model significantly outperforms various baselines on three datasets, demonstrating its superiority for the task.
\end{abstract}

\section{Introduction}
Discourse relations, such as \textit{Cause} and \textit{Contrast}, describe the logical relation between two text spans ~\citep{pitler-etal-2009-automatic}. Recognizing discourse relations is beneficial for various NLP tasks, including coherence modeling ~\citep{lin-etal-2011-automatically}, reading comprehension ~\citep{mihaylov-frank-2019-discourse}, argumentation mining ~\citep{habernal-gurevych-2017-argumentation,hewett-etal-2019-utility}, and machine translation ~\citep{meyer2015discourse,longyue2019discourse}. 

Discourse connectives (e.g., \textit{but}, \textit{as a result}) are words or phrases that signal the presence of a discourse relation ~\citep{pitler-nenkova-2009-using}. They can be explicit, as in (1), or implicit, as in (2):
\begin{itemize}
    \item[(1)] [I refused to pay the cobbler the full \$95]\textsubscript{Arg1} \underline{\textbf{because}} [he did poor work.]\textsubscript{Arg2}

    \item[(2)] [They put the treasury secretary back on the board.]\textsubscript{Arg1} (\underline{\textbf{Implicit=However}}) [There is doubt that the change would accomplish much.]\textsubscript{Arg2}
\end{itemize}
When discourse connectives are explicitly present between arguments, classifying the sense of a discourse relation is straightforward. For example, ~\citet{pitler-nenkova-2009-using} proved that using only connectives in a text as features, the accuracy of 4-way explicit discourse relation classification on PDTB 2.0 can reach 85.8\%. However, for implicit cases, there are no connectives to explicitly mark discourse relations, which makes implicit discourse relation classification challenging ~\citep{zhou-etal-2010-effects,shi-etal-2017-using}. Existing work attempts to perform implicit discourse relation classification directly from arguments. They range from designing linguistically informed features from arguments ~\cite{lin-etal-2009-recognizing,pitler-etal-2009-automatic} to modeling interaction between arguments using neural networks ~\citep{ijcai-lei,guo-etal-2018-implicit}. Despite their impressive performance, the absence of explicit discourse connectives makes the prediction extremely hard and hinders further improvement ~\citep{lin2014pdtb,qin-etal-2017-adversarial}. 

The huge performance gap between explicit and implicit classification (85.8\% vs. 57.6\%) ~\citep{liu-li-2016-recognizing} motivates recent studies to utilize implicit connectives for the training process of implicit relation classifiers. For instance, ~\citet{qin-etal-2017-adversarial} developed an adversarial model to transfer knowledge from the model supplied with implicit connectives to the model without such information, while \citet{kishimoto-etal-2020-adapting} proposed a multi-task learning framework to incorporate implicit connectives prediction as another training objective. However, we argue that these methods are suboptimal since connectives are still not explicitly present in input texts. This is demonstrated by \citet{kishimoto-etal-2020-adapting}, concluding that adding implicit connective prediction as a training objective provides only negligible gain for implicit relation classification on PDTB 2.0 (we empirically found the conclusion also held on the adversarial model). 

In this paper, we design a novel end-to-end model to leverage discourse connectives for the task of implicit discourse relation classification. The key inspiration is derived from the annotation process of implicit discourse relations in PDTB, which consists of inserting a connective that best conveys the inferred relation, and annotating the relation label based on both the inserted implicit connectives and contextual semantics ~\citep{prasad-etal-2008-penn}. We imitate this process by explicitly gene\-rating discourse connectives for the implicit relation class\-ifier. Specifically, our model jointly learns to gene\-rate discourse connectives between arguments and predict discourse relations based on the arguments and the generated connectives. A potential drawback of this joint model is that the poorly generated connectives at the early stage of joint training may mislead the relation classifier. One possible solution is always feeding true connectives to the implicit relation classifier for training. But it leads to severe discrepancies between training and inference ~\citep{sporleder_lascarides_2008}, since manually-annotated connectives are unavailable during evaluation ~\citep{prasad-etal-2008-penn}. To address this issue, we adopt Scheduled Sampling ~\citep{bengio2015scheduled} into our method. To be more specific, our relation classifier is first trained with hand-annotated implicit connectives and then gradually shifts to use generated connectives.

We evaluate our model\footnote{\url{https://github.com/liuwei1206/ConnRel}} on two English corpora\-, PDTB 2.0 ~\citep{prasad-etal-2008-penn}, PDTB 3.0 ~\citep{webber2019penn}, and a German corpus, PCC ~\citep{bourgonje-stede-2020-potsdam}, and compare it with other connective-enhanced approaches and existing state-of-the-art works. Results show that our method significantly outperforms those connective-enhanced baselines on three datasets while offering comparable performance to existing sota models. 

In addition, we perform the first systematic analysis of different connective-enhanced models to investigate why our method works better. Our studies show that: (1) models learn to use connectives more effectively when putting connectives in the input rather than using them as training objectives; (2) end-to-end training can improve models' robustness to incorrectly-predicted connectives; (3) our method shows a better balance between arguments and connectives for relation prediction than other baselines. Finally, we show that connectives can effectively improve the predictive performance on frequent relations while failing on those with limited training instances.

\section{Related Work}
Implicit discourse relation classification, as a challenging part of shallow discourse parsing, has drawn much attention since the release of PDTB 2.0 ~\citep{prasad-etal-2008-penn}. Most of the work focused on predicting implicit relations directly from input arguments. For example, early statistical methods have put much effort into designing linguistically informed features from arguments ~\citep{pitler-etal-2009-automatic,pitler-nenkova-2009-using,lin-etal-2009-recognizing,rutherford-xue-2014-discovering}. More recently, neural networks ~\citep{zhang-etal-2015-shallow,kishimoto-etal-2018-knowledge, BMGF, LDSGM, long-webber-2022-facilitating} have been applied to learning useful semantic and syntactic information from arguments due to their strength in representation learning. Despite achieving impressive results, the absence of connectives makes their performance still lag far behind explicit discourse parsing. 

\begin{figure*}[t!]
\centering\includegraphics[scale=0.445,trim=0 0 0 0]{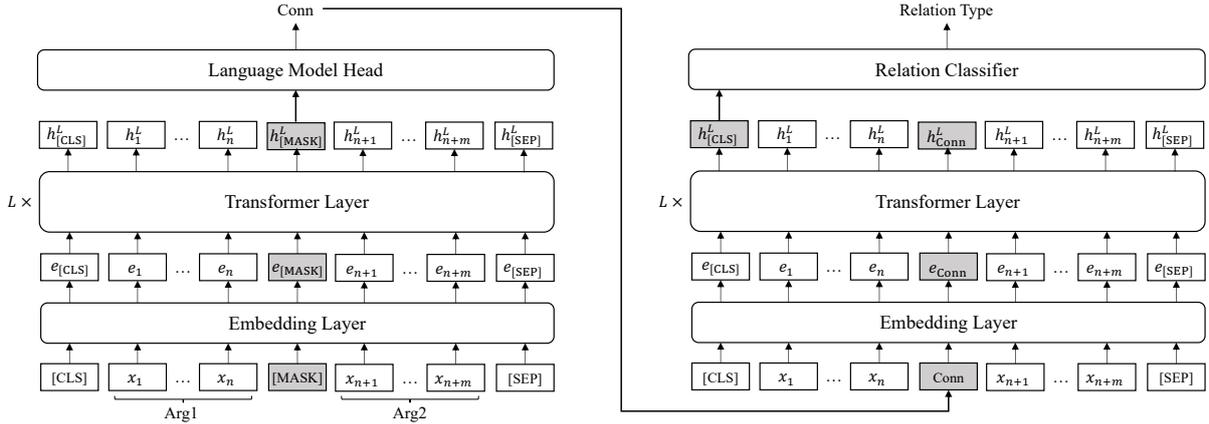}
\setlength{\abovecaptionskip}{-5pt}
\setlength{\belowcaptionskip}{-12pt}
\caption{An overview of the proposed approach. The left part is the connective generation module which generates a connective at the masked position between arguments (Arg1, Arg2). The right part is the relation classification module which predicts the relation based on both arguments and the generated connective. We share the embedding layer and transformer blocks between two modules and train the whole model in an end-to-end manner.}
\label{figure:framework}
\end{figure*}

The question of how to leverage discourse connectives for implicit discourse relation classification has received continued research attention. ~\citet{zhou-etal-2010-effects} proposed a pipeline method to investigate the benefits of connectives recovered from an n-gram language model for implicit relation recognition. Their results show that using recovered connectives as features can achieve comparable performance to a strong baseline. This pipeline-based method is further improved by following efforts, including integrating pre-trained models ~\citep{kurfali-ostling-2021-lets, pipejiang} and using prompt strategies ~\citep{xiang-etal-2022-connprompt, zhou-etal-2022-prompt-based}. However, some works ~\citep{qin-etal-2017-adversarial,xiang2022survey} pointed out that pipeline methods suffer cascading errors. Recent studies have shifted to using end-to-end neural networks. ~\citet{qin-etal-2017-adversarial} proposed a feature imitation framework in which an implicit relation network is driven to learn from another neural network with access to connectives. ~\citet{shi-demberg-2019-learning} designed an encoder-decoder model that generates implicit connectives from texts and learns a relation classifier using the representation of the encoder. ~\citet{kishimoto-etal-2020-adapting} investigated a multi-task learning approach to predict connectives and discourse relations simultaneously. Our method is in line with those recent approaches exploiting connectives with an end-to-end neural network. The main difference is that those models focus on using implicit connectives in a non-input manner (i.e. they do not input implicit connectives as features but utilize them as another training signal), whereas our method explicitly generates connectives and inputs both arguments and the generated connectives into the relation classifier.

Our method can be viewed as a joint learning framework. Such a framework has been used to learn information exchange and reduce error propagation between related tasks ~\citep{zhang-2018-joint}. ~\citet{collobert2011natural} designed a unified neural model to perform tagging, chunking, and NER jointly. ~\citet{sogaard-goldberg-2016-deep} refined this unified framework by putting low-level tasks supervised at lower layers. ~\citet{miwa-bansal-2016-end} presented an LSTM-based model to extract entities and the relations between them. ~\citet{strubell-etal-2018-linguistically} proposed a joint model for semantic role labeling (SRL), in which dependency parsing results were used to guide the attention module in the SRL task. Compared with these works, our joint learning framework is different in both motivation and design. For example, instead of simply sharing an encoder between tasks, we input the results of connective generation into the relation classifier.

\section{Method}
Inspired by the annotation process of PDTB, we explicitly generate discourse connectives for implicit relation classification. Following previous work ~\citep{lin-etal-2009-recognizing}, we use the gold standard arguments and focus on relation prediction. Figure \ref{figure:framework} shows the overall architecture of our proposed model. It consists of two components: (1) generating a discourse connective between arguments; (2) predicting discourse relation based on arguments and the generated connective. In this section, we describe these two components in detail and show the challenges during training and our solutions.

Formally, let $X_1=\{x_1, ..., x_n\}$ and $X_2=\{x_{n+1}, ..., x_{n+m}\}$ be the two input arguments (Arg1 and Arg2) of implicit relation classification, where $x_i$ denotes the $i$-th word in Arg1 and $x_{n+j}$ denotes the $j$-th word in Arg2. We denote the relation between those two arguments as $y$. Similar to the setup in existing connective enhanced methods, each training sample $(X_1, X_2, c, y)$ also includes an annotated implicit connective $c$ that best expresses the relation. During the evaluation, only arguments $(X_1, X_2)$ are available to the model.

\subsection{Connective Generation}
Connective generation aims to generate a discourse connective between two arguments (shown in the left part of Figure \ref{figure:framework}). We achieve this by using bidirectional masked language models ~\citep{devlin-etal-2019-bert}, such as RoBERTa. Specifically, we insert a $\rm{[MASK]}$ token between two arguments and gene\-rate a connective on the masked position. 

Given a pair of arguments Arg1 and Arg2, 
we first concatenate a $\rm{[CLS]}$ token, argument Arg1, a $\rm{[MASK]}$ token, argument Arg2, and a $\rm{[SEP]}$ token into $\widetilde{X} = \{{\rm [CLS]} \; X_1 \; {\rm [MASK]} \; X_2 \; {\rm [SEP]}\}$. For each token $\tilde{x}_i$ in $\widetilde{X}$, we convert it into the vector space by adding token, segment, and position embeddings, thus yielding input embeddings $E \in {\mathbb{R}}^{(n+m+3)\times d}$, where $d$ is the hidden size. Then we input $E$ into $L$ stacked Transformer blocks, and each Transformer layer acts as follows:
\begin{equation}
    \begin{aligned}
        G & = {\rm LN}(H^{l-1} + {\rm MHAttn}(H^{l-1})) \\
        H^{l} & = {\rm LN}(G + {\rm FFN}(G)) \\
    \end{aligned}
    \vspace{-5pt}
\end{equation}
where $H^l$ denotes the output of the $l$-th layer and $H^0=E$; $\rm LN$ is layer normalization; $\rm MHAttn$ is the multi-head attention mechanism; $\rm FFN$ is a two-layer feed-forward network with $\rm ReLU$ as hidden activation function. To generate a connective on the masked position, we feed the hidden state of the $\rm [MASK]$ token after $L$ Transformer layers into a language model head ($\rm LMHead$):
\begin{equation}
    \setlength{\abovedisplayskip}{6pt}
    \setlength{\belowdisplayskip}{6pt}
    \mathbf{p}^c = {\rm LMHead}(h_{{\rm [MASK]}}^{L})
\end{equation}
where $\mathbf{p}^c$ denotes the probabilities over the whole connective vocabulary. However, a normal $\rm LMHead$ can only generate one word without the capacity to generate multi-word connectives, such as "for instance". To overcome this shortcoming, we create several special tokens in $\rm LMHead$'s vocabulary to represent those multi-word connectives, and initialize their embedding with the average embedding of the contained single words. Taking "for instance" as an example, we create a token $\rm [for\_instance]$ and set its embedding as ${\rm Average}({\rm embed}(\text{"for"}), {\rm embed}(\text{"instance"}))$.

We choose cross-entropy as loss function for the connective generation module:
\begin{equation}
    \setlength{\abovedisplayskip}{6pt}
    \setlength{\belowdisplayskip}{6pt}
    \mathcal{L}_{Conn} = -\sum_{i=0}^{N} \sum_{j=0}^{CN} C_{ij} \log (P_{ij}^c)
\end{equation}
where $C_{i}$ is the annotated implicit connective of the $i$-th sample with a one-hot scheme, $CN$ is the total number of connectives. 

\subsection{Relation Classification}
The goal of relation classification is to predict the implicit relation between arguments. Typically, it is solved using only arguments as input ~\citep{zhang-etal-2015-shallow,kishimoto-etal-2018-knowledge}. In this work, we propose to predict implicit relations based on both input arguments and the generated connectives (shown in the right part of Figure \ref{figure:framework}). 

First, we need to obtain a connective from the connective generation module. A straightforward way to do so is to apply the ${\rm arg\;max}$ operation on the probabilities output by $\rm LMHead$, i.e. $\rm Conn$ = ${\rm arg\;max}(\mathbf{p}^c)$. However, it is a non-differentiable process, which means the training signal of relation classification can not be propagated back to adjust the parameters of the connective generation module. Hence, we adopt the Gumbel-Softmax technique ~\citep{gumbel} for the task. The Gumbel-Softmax technique has been shown to be an effective approximation to the discrete variable ~\citep{shi-etal-2021-neural}. Therefore, we use
\begin{equation}
    \setlength{\abovedisplayskip}{6pt}
    \setlength{\belowdisplayskip}{6pt}
    \label{equ:gumbel}
    \begin{aligned}
        g &= -\log(-\log(\xi)), \; \xi \sim {\rm U}(0,1) \\
        \mathbf{c}_i &= \frac{\exp((\log(p_i^c)+g_i)/\tau)}{\sum_j \exp((\log(p_j^c)+g_j)/\tau)}
    \end{aligned}
\end{equation}
as the approximation of the one-hot vector of the generated connective on the masked position (denoted as $\rm Conn$ in Figure \ref{figure:framework}), where $g$ is the Gumbel distribution, $\rm U$ is the uniform distribution, $p_i^c$ is the probability of $i$-th connective output by the $\rm LMHead$, $\tau \in (0, \infty)$ is a temperature parameter. 

After we have obtained the generated connective "$\rm Conn$", we concatenate it with arguments and construct a new input as $\bar{X}=\{{\rm [CLS]}\;X_1\;\rm{Conn}\;X_2\;{\rm [SEP]}\}$. This new form of input is precisely the same as the input in explicit discourse relation classification. We argue that the key to fully using connectives is to insert them into the input texts instead of treating them simply as a training objective. Like the connective generation module, we feed $\bar{X}$ into an Embedding layer and $L$ stacked Transformer blocks. Note that we share the Embedding Layer and Transformers between connective generation and relation classification modules. Doing so can not only reduce the total memory for training the model but also prompt the interaction between two tasks. Finally, we feed the outputs of the $L$-th Transformer at $\rm [CLS]$ position to a relation classification layer:
\begin{equation}
    \setlength{\abovedisplayskip}{5pt}
    \setlength{\belowdisplayskip}{5pt}
    \mathbf{p}^r = {\rm softmax}(\mathbf{W}_r h_{\rm [CLS]}^L+\mathbf{b}_r)
\end{equation}
where $\mathbf{W}_r$ and $\mathbf{b}_r$ are learnable parameters. Similarly, we use cross-entropy for training, and the loss is formulated as:
\begin{equation}
    \setlength{\abovedisplayskip}{6pt}
    \setlength{\belowdisplayskip}{6pt}
    \mathcal{L}_{Rel} = -\sum_{i=0}^{N} \sum_{j=0}^{RN} Y_{ij} \log (P_{ij}^r)
\end{equation}
where $Y_{i}$ is the ground truth relation of the $i$-th sample with a one-hot scheme, $RN$ is the total number of relations.

\subsection{Training and Evaluation}
To jointly train those two modules, we use a multi-task loss:
\begin{equation}
    \setlength{\abovedisplayskip}{5pt}
    \setlength{\belowdisplayskip}{5pt}
    \mathcal{L} = \mathcal{L}_{Conn} + \mathcal{L}_{Rel}
\end{equation}
A potential issue of this training is that poorly generated connectives at an early stage of joint training may mislead the relation classifier. One possible solution is always providing manually annotated implicit connectives to the relation classifier, similar to Teacher Forcing ~\citep{RanzatoCAZ15}. But this might lead to a severe discrepancy between training and inference since manually annotated connectives are not available during inference. We solve those issues by introducing Scheduled Sampling ~\citep{bengio2015scheduled} into our method. Scheduled Sampling is designed to sample tokens between gold references and model predictions with a scheduled probability in seq2seq models. We adopt it into our training by sampling between manually-annotated and the generated connect\-ives. Specific\-ally, we use the inverse sigmoid decay ~\citep{bengio2015scheduled}, in which probability of sampling manually annotated connectives at the $t$-th training step is calculated as follows:
\begin{equation}
    \setlength{\abovedisplayskip}{6pt}
    \setlength{\belowdisplayskip}{6pt}
    \label{equ:scheduled}
    \epsilon_t = \frac{k}{k + \exp(t/k)}
\end{equation}
where $k \geq 1$ is a hyper-parameter to control the convergence speed. In the beginning, training is similar to Teacher Forcing due to $\epsilon_t \approx 1$. As the training step $t$ increases, the relation classifier gradually uses more generated connectives, and eventually uses only generated ones (identical to the evaluation setting) when $\epsilon_t \approx 0$. We show the sampling process during training in Algorithm \ref{al:ss}. 

\setlength{\textfloatsep}{6pt}
\begin{algorithm}[t]
	\renewcommand{\algorithmicrequire}{\textbf{Input:}}
	\renewcommand{\algorithmicensure}{\textbf{Output:}}
	\caption{Scheduled Sampling in Training}
	\begin{algorithmic}[1]
	    \Require
	    	relation classifier $\rm \mathbf{RelCls}$, arguments $\rm X_1, X_2$, annotated connective $\rm true\_conn$, gene\-rated connective $\rm gene\_conn$, training step $\rm t$, hyper\-parameter in decay $\rm k$ 
	    \Ensure
	        $\rm logits$
	    \State $\rm p = random()$ \Comment{$[0.0, 1.0)$}
	    \State $\rm \epsilon_t = \frac{k}{k + \exp(t/k)}$
	    \If{$\rm p < \epsilon_t $}
		    \State $\rm logits = \mathbf{RelCls}(X_1, X_2, true\_conn)$
		\Else
			\State $\rm logits = \mathbf{RelCls}(X_1, X_2, gene\_conn)$
		\EndIf
	\end{algorithmic}  
	\label{al:ss}
\end{algorithm}

During inference, we generate a connective $\rm Conn$ through ${\rm arg\;max (\mathbf{p^c}})$, feed the generated $\rm Conn$ and arguments into the relation classifier, and choose the relation type that possesses the maximum value in $\mathbf{p}^r$.

\section{Experiments}
We carry out a set of experiments to investigate the effectiveness of our method across different corpora and dataset splittings. In addition, we perform analyses showing that our model learns a better balance between using connectives and arguments than baselines.

\subsection{Experimental Settings}
\textbf{Datasets.} We evaluate our model on two English corpora, PDTB 2.0 ~\citep{prasad-etal-2008-penn}, PDTB 3.0 ~\citep{webber2019penn}, and a German corpus, PCC ~\citep{bourgonje-stede-2020-potsdam}. In PDTB, instances are annotated with senses from a three-level sense hierarchy. We follow previous works ~\citep{ji-eisenstein-2015-one,kim-etal-2020-implicit} to use top-level 4-way and second-level 11-way classification for PDTB 2.0, and top-level 4-way and second-level 14-way for PDTB 3.0. As for the dataset split, we adopt two different settings for both PDTB 2.0 and PDTB 3.0. The first one is proposed by ~\citet{ji-eisenstein-2015-one}, where sections 2-20, sections 0-1, and sections 21-22 are used as training, development, and test set. The second one is called section-level cross-validation ~\citep{kim-etal-2020-implicit}, in which 25 sections are divided into 12 folds with 2 validation, 2 test, and 21 training sections. There are over one hundred connectives in PDTB (e.g., 102 in PDTB 2.0), but some rarely occur (e.g., only 7 for "next" in PDTB 2.0). To reduce the complexity of connective generation and ensure each connective has sufficient training data, we only consider connectives with a frequency of at least 100 in the experiments. PCC is a German corpus following the annotation guidelines of PDTB. For this corpus, we only use the second-level 8-way classification since the distribution of top-level relations is highly uneven ~\citep{bourgonje:diss}. A more detailed description and statistics of the datasets are given in Appendix \ref{sec:data}.

\begin{table*}[t]
\centering
\scalebox{0.85}{
\begin{tabular}{l|cccc|cccc}
\hline
            & \multicolumn{4}{c|}{Level1 4-way}                  & \multicolumn{4}{c}{Level2 11-way}                 \\
            & \multicolumn{2}{c}{Ji}\hspace{1.3em} & \multicolumn{2}{c}{Xval}\hspace{1.3em} & \multicolumn{2}{|c}{Ji}\hspace{1.3em} & \multicolumn{2}{c}{Xval}\hspace{1.3em} \\ \hline
Models      & Acc\hspace{1.3em}        & F1\hspace{1.3em}        & Acc\hspace{1.3em}         & F1\hspace{1.3em}          & Acc\hspace{1.3em}        & F1\hspace{1.3em}        & Acc\hspace{1.3em}         & F1\hspace{1.3em}         \\ \hline
~\citet{BMGF}      & 69.06\textsubscript{0.43}        & 63.39\textsubscript{0.56}         & -\hspace{1.3em}          & -\hspace{1.3em}         & 58.13\textsubscript{0.67}          & -\hspace{1.3em}        & -\hspace{1.3em}         & -\hspace{1.3em} \\
~\citet{kim-etal-2020-implicit}      & 66.30\hspace{1.3em}        &56.00\hspace{1.3em}         & -\hspace{1.3em}          & -\hspace{1.3em}         & 54.73\textsubscript{0.79}          & -\hspace{1.3em}        & 52.98\textsubscript{0.29}         & -\hspace{1.3em} \\ 
~\citet{LDSGM}      & 71.18\hspace{1.3em}        & 63.73\hspace{1.3em}         & -\hspace{1.3em}          & -\hspace{1.3em}         & 60.33\hspace{1.3em}          & 40.49\hspace{1.3em}        & -\hspace{1.3em}         & -\hspace{1.3em} \\
~\citet{zhou-etal-2022-prompt-based}      & 70.84\hspace{1.3em}        & 64.95\hspace{1.3em}         & -\hspace{1.3em}          & -\hspace{1.3em}         & 60.54\hspace{1.3em}          & 41.55\hspace{1.3em}        & -\hspace{1.3em}         & -\hspace{1.3em} \\
~\citet{long-webber-2022-facilitating}      & 72.18\hspace{1.3em}        & \textbf{69.60}\hspace{1.3em}         & -\hspace{1.3em}          & -\hspace{1.3em}         & 61.69\hspace{1.3em}          & \textbf{49.66}\hspace{1.3em}        & -\hspace{1.3em}         & -\hspace{1.3em} \\ \hline
RoBERTa     & 68.61\textsubscript{0.73}      & 60.89\textsubscript{0.19}     & 68.66\textsubscript{1.29}       & 60.49\textsubscript{1.86}       & 58.84\textsubscript{0.48}      & 39.31\textsubscript{0.83}     & 55.40\textsubscript{1.65}       & 36.51\textsubscript{2.75}      \\
RoBERTaConn & 55.34\textsubscript{0.39}      & 37.47\textsubscript{2.27}     & 54.28\textsubscript{2.12}       & 34.71\textsubscript{2.75}       & 31.97\textsubscript{2.75}      & 17.10\textsubscript{2.81}     & 32.12\textsubscript{2.63}       & 17.91\textsubscript{2.12}      \\
Adversarial & 69.43\textsubscript{0.70}      & 62.44\textsubscript{0.61}     & 69.13\textsubscript{1.14}       & 60.63\textsubscript{1.47}       & 57.63\textsubscript{1.10}      & 38.81\textsubscript{2.25}     & 54.43\textsubscript{1.79}       & 36.79\textsubscript{2.24}      \\
Multi-Task  & 70.82\textsubscript{0.72}      & 63.79\textsubscript{0.82}     & 70.02\textsubscript{1.40}       & 62.19\textsubscript{1.84}       & 60.21\textsubscript{0.94}      & 39.75\textsubscript{0.70}     & 56.85\textsubscript{1.13}       & 36.83\textsubscript{2.42}      \\
Pipeline    & 71.01\textsubscript{0.89}      & 64.65\textsubscript{1.03}     & 69.12\textsubscript{1.03}       & 61.65\textsubscript{0.89}       & 59.42\textsubscript{0.54}      & 40.84\textsubscript{0.39}     & 55.24\textsubscript{1.72}       & 37.03\textsubscript{2.83}      \\
\hline
Our Model   & \textbf{74.59}\textsubscript{0.44}      & 68.64\textsubscript{0.67}     & \textbf{71.33}\textsubscript{1.25}       & \textbf{63.84}\textsubscript{1.96}       & \textbf{62.75}\textsubscript{0.59}      & 42.36\textsubscript{0.38}     & \textbf{57.98}\textsubscript{1.22}       & \textbf{39.05}\textsubscript{3.53}      \\ \hline
\end{tabular}}
\setlength{\abovecaptionskip}{10pt}
\setlength{\belowcaptionskip}{-8pt} 
\caption{Results on PDTB \textbf{2.0}. Subscripts are the standard deviation of the mean performance.}
\label{table:pdtb2}
\end{table*}

\noindent \textbf{Implementation Details.} We implement our model using the Pytorch library. The bidirectional masked language model used in our work is $\rm RoBERTa_{base}$, which is initialized with the pre-trained checkpoint from Huggingface. For hyperparameter configurations, we mainly follow the settings in RoBERTa ~\citep{roberta}. We use the AdamW optimizer with an initial learning rate of 1e-5, a batch size of 16, and a maximum epoch number of 10 for training. Considering the training variability in PDTB, we report the mean performance of 5 random restarts for the "Ji" splits and that of the section-level cross-validation (Xval) like ~\citet{kim-etal-2020-implicit}. For PCC, we conduct a 5-fold cross-validation (Xval) on this corpus due to its limited number of data. We use standard accuracy (Acc, \%) and F1-macro (F1, \%) as evaluation metrics. We show more detailed settings and hyperparameters in Appendix \ref{sec:setting}. 

\noindent \textbf{Baselines.} To demonstrate the effectiveness of our model, we compare it with state-of-the-art connective-enhanced methods and several variants of our model:
\begin{itemize}[leftmargin=*]
    \setlength\itemsep{-0.2em}
    \item \textbf{RoBERTa}. Finetune RoBERTa for implicit relation classification. Only arguments (Arg1, Arg2) are input for training without using any implicit discourse connective information.
    \item \textbf{RoBERTaConn}. A variant of the RoBERTa baseline. During training, we feed both arguments and annotated connectives, i.e., (Arg1, Arg2, true\_conn), to RoBERTa. During inference, only arguments (Arg1, Arg2) are input to the model.
    \item \textbf{Adversarial}. An adversarial-based connective enhanced method ~\citep{qin-etal-2017-adversarial}, in which an implicit relation network is driven to learn from another neural network with access to connectives. We replace its encoder with $\rm RoBERTa_{base}$ for a fair comparison.
    \item \textbf{Multi-Task}. A multi-task framework for implicit relation classification ~\citep{kishimoto-etal-2020-adapting}, in which connective prediction is introduced as another training task. We equip it with the same $\rm RoBERTa_{base}$ as our method.
    \item \textbf{Pipeline}. A pipeline variant of our method, in which we first train a connective generation model, then learn a relation classifier with arguments and the generated connectives. Note that these two modules are trained separately.
\end{itemize}
Further, we compare our method against previous state-of-the-art models on each corpus.

\begin{table*}[t]
\centering
\scalebox{0.71}{
\begin{tabular}{l|cccc|cccc|cc}
\hline
            & \multicolumn{4}{c|}{Level1 4-way (\textbf{PDTB 3.0})}                  & \multicolumn{4}{c|}{Level2 14-way (\textbf{PDTB 3.0})}  & \multicolumn{2}{c}{Level2 8-way (\textbf{PCC})}                \\
            & \multicolumn{2}{c}{Ji}\hspace{1.3em} & \multicolumn{2}{c}{Xval}\hspace{1.3em} & \multicolumn{2}{|c}{Ji}\hspace{1.3em} & \multicolumn{2}{c}{Xval}\hspace{1.3em} 
             & \multicolumn{2}{|c}{Xval}\hspace{1.3em}  \\ \hline
Models      & Acc\hspace{1.25em}        & F1\hspace{1.25em}        & Acc\hspace{1.25em}         & F1\hspace{1.25em}          & Acc\hspace{1.25em}        & F1\hspace{1.25em}        & Acc\hspace{1.25em}         & F1\hspace{1.25em}   & Acc\hspace{1.25em} & F1\hspace{1.25em}      \\ \hline
~\citet{kim-etal-2020-implicit}      & 71.30\hspace{1.3em}        & 64.80\hspace{1.3em}         & -\hspace{1.25em}          & -\hspace{1.25em}         & -\hspace{1.25em}          & -\hspace{1.25em}        & 60.78\textsubscript{0.24}         & -\hspace{1.25em}  & -\hspace{1.25em} & -\hspace{1.25em} \\
~\citet{xiang-etal-2022-connprompt}      & 74.36\hspace{1.3em}        &69.91\hspace{1.3em}         & -\hspace{1.25em}          & -\hspace{1.25em}         & -\hspace{1.25em}          & -\hspace{1.25em}        & -\hspace{1.25em}         & -\hspace{1.25em} & -\hspace{1.25em} & -\hspace{1.25em} \\
~\citet{long-webber-2022-facilitating}      & 75.31\hspace{1.3em}        &70.05\hspace{1.3em}         & -\hspace{1.25em}          & -\hspace{1.25em}         & 64.68\hspace{1.3em}          & \textbf{57.62}\hspace{1.3em}        & -\hspace{1.25em}         & -\hspace{1.25em} & -\hspace{1.25em} & -\hspace{1.25em} \\ \hline
RoBERTa     & 73.51\textsubscript{0.69}      & 67.98\textsubscript{0.97}     & 73.42\textsubscript{0.90}       & 67.54\textsubscript{1.40}       & 63.32\textsubscript{0.40}      & 52.49\textsubscript{1.26}     & 62.65\textsubscript{1.32}       & 53.19\textsubscript{1.20} & 35.80\textsubscript{1.13} & 15.08\textsubscript{0.97}      \\
RoBERTaConn & 51.74\textsubscript{0.76}      & 41.45\textsubscript{0.69}     & 53.90\textsubscript{1.71}       & 39.39\textsubscript{2.74}       & 33.67\textsubscript{1.78}      & 25.40\textsubscript{2.11}     & 36.68\textsubscript{2.39}       & 28.18\textsubscript{4.11} & 30.30\textsubscript{2.86} & 12.62\textsubscript{2.06}     \\
Adversarial & 73.83\textsubscript{0.28}      & 68.60\textsubscript{0.75}     & 73.30\textsubscript{1.32}       & 67.23\textsubscript{1.85}       & 63.00\textsubscript{0.48}      & 54.28\textsubscript{1.76}     & 62.12\textsubscript{1.46}       & 53.85\textsubscript{1.46} & 35.02\textsubscript{3.18} & 18.48\textsubscript{1.51}     \\
Multi-Task  & 74.97\textsubscript{0.70}      & 69.67\textsubscript{0.76}     & 73.83\textsubscript{0.94}       & 68.04\textsubscript{1.30}       & 64.52\textsubscript{0.31}      & 53.12\textsubscript{0.63}     & 62.81\textsubscript{1.36}       & 53.07\textsubscript{1.40} & 40.48\textsubscript{1.47} & 21.22\textsubscript{2.01}      \\
Pipeline    & 74.54\textsubscript{0.22}      & 69.19\textsubscript{0.60}     & 73.70\textsubscript{0.89}       & 68.31\textsubscript{1.78}       & 63.98\textsubscript{0.63}      & 52.95\textsubscript{0.48}     & 63.07\textsubscript{1.70}       & 53.43\textsubscript{1.63} & 42.97\textsubscript{3.48} & 22.66\textsubscript{1.20}      \\ \hline
Our Model   & \textbf{76.23}\textsubscript{0.19}      & \textbf{71.15}\textsubscript{0.47}     & \textbf{75.41}\textsubscript{0.89}       & \textbf{70.06}\textsubscript{1.72}       & \textbf{65.51}\textsubscript{0.41}      & 54.92\textsubscript{0.81}     & \textbf{64.59}\textsubscript{1.21}       & \textbf{55.26}\textsubscript{1.32} & \textbf{44.54}\textsubscript{3.06} & \textbf{26.93}\textsubscript{2.06}     \\ \hline
\end{tabular}}
\setlength{\abovecaptionskip}{8pt}
\setlength{\belowcaptionskip}{-11pt} 
\caption{Results on PDTB \textbf{3.0} and \textbf{PCC}. Subscripts are the standard deviation of the mean performance.}
\label{table:pdtb3}
\end{table*}

\subsection{Overall Results}
\noindent \textbf{PDTB 2.0}. Table \ref{table:pdtb2} shows the experimental results on PDTB 2.0. RoBERTaConn shows a much worse performance than the RoBERTa baseline on this corpus, indicating that simply feeding annotated connectives to the model causes a severe discrepancy between training and evaluation. This is also somewhat in accord with ~\citet{sporleder_lascarides_2008}, which shows that models trained on explicitly-marked examples generalize poorly to implicit relation identification. Discourse connective-enhanced models, including Adversarial\-, Multi-Task, Pipeline and Our Method, achieve better perform\-ance than the RoBERTa baseline. This demonstrates that utilizing the annotated connectives information for training is beneficial for implicit relation classification. The improvement of Adversarial and Multi-task over the RoBERTa baseline is limited and unstable. We argue this is because they do not exploit connectives in the way of input features but treat them as training objectives, thus limiting connectives' contributions to implicit relation classification. Pipeline also shows limited performance gain over the baseline. We speculate that this is due to its pipeline setting (i.e. connective generation $\to$ relation classification), which propagates errors in connective generation to relation classification ~\citep{qin-etal-2017-adversarial}. Compared to the above connective-enhanced models, our method's improvement over the RoBERTa baseline is bigger, which suggests that our approach is more efficient in utilizing connectives. To further show the efficiency\- of our method, we compare it against previous state-of-the-art models on PDTB 2.0 ~\citep{BMGF, kim-etal-2020-implicit, LDSGM, zhou-etal-2022-prompt-based, long-webber-2022-facilitating}. The first block of Table \ref{table:pdtb2} shows the results of those models, from which we observe that our model outperforms most of them, especially on accuracy, achieving the best results on this corpus. The only exception is that the F1-score of our method lags behind ~\citet{long-webber-2022-facilitating}, particularly on level2 classification. This is because our method cannot predict several fine-grained relations (see Section \ref{sec-rel_ana}), such as Comparison.Concession, which leads to the low averaged F1 at the label-level. 

\noindent \textbf{PDTB 3.0 / PCC}. Results on PDTB 3.0 and PCC are shown in Table \ref{table:pdtb3}. Similar to the results on the PDTB 2.0 corpus, simply feeding connectives for training (RoBERTaCon\-n) hurts the performance, especially on the Level2 classification of PDTB 3.0. Adversarial and Multi-Task perform better than the RoBERTa baseline, although their improvement is limited. Despite suffering cascading errors, Pipeline shows comparative and even better results than Adversarial and Multi-Task on the two corpora. This indicates the advantage of utilizing connectives as input features rather than a training objective, particularly on PCC. Consistent with the results on PDTB 2.0, our method outperforms Adversarial, Multi-task, and Pipeline on both datasets, demonstrating the superiority of inputting connect\-ives to the relation classifier in an end-to-end manner and also showing that it works well on different languages. We further compare our method with three existing sota models on PDTB 3.0, ~\citet{kim-etal-2020-implicit}, ~\citet{xiang-etal-2022-connprompt}, and ~\citet{long-webber-2022-facilitating}. Results in Table \ref{table:pdtb3} show that our approach performs better than these three models.

\subsection{Performance Analysis}
\label{sec:detail_ana}
To figure out why our model works well, we first perform analyses on its behavior answering two questions: (1) whether it really benefits from discourse connectives; (2) whether it can also make correct predictions when connectives are missing. We then investigate the relation classifier's performance in the different models when connectives are correctly and incorrectly generated (or predicted). 

\begin{figure}[t]
\centering
\includegraphics[scale=0.47,trim=0 0 0 0]{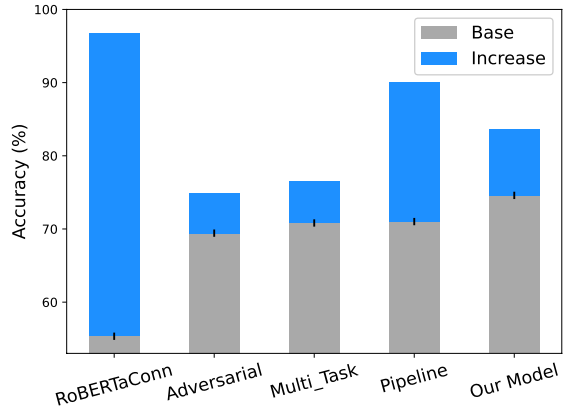}
\setlength{\abovecaptionskip}{6pt} 
\caption{Level1 classification results on PDTB 2.0 (Ji split) when annotated connectives are fed to connective-enhanced models. "Increase" denotes performance gain compared to the model with default settings ("Base").}
\label{figure:pdtb2_l1_acc_inc}
\end{figure}

We perform the first analysis by replacing the generated connectives in our model with manually-annotated ones\footnote{In PDTB 2.0 and PDTB 3.0, each instance contains annotated implicit connectives, making this analysis possible.\label{foot:gt_conn}}, and compare its performance before and after this setup. Intuitively, if our model benefits from discourse connectives, accuracy and F1-macro should increase after the change. For comparison, we apply the same setup to other connective-enhanced models. We conduct experiments\footnote{We show more detailed results and also case studies in Appendix \ref{app:sec:detailed_behavior}.\label{foot:detail_ana}} on the Level1 classification of PDTB 2.0 (Ji split), and show the accuracy results in Figure \ref{figure:pdtb2_l1_acc_inc}. As expected, our model's performance shows a substantial improvement, demonstrating that it does learn to use discourse connectives for implicit relation classification. Other connective-enhanced models also perform better in such a setup but with a different degree of gain. Specifically, models that use connectives as input features during training (RoBERTaConn, Pipeline, and Our Method) show more increase and have higher upper bounds than models that use connectives as training objectives (Adversarial and Multi-Task). This aligns with our assumption that putting connectives in the input is more efficient for a model learning to use discourse connectives for implicit relation classification than treating them as training objectives. However, inputting connectives for training can lead to another severe issue, i.e., the model relies too much on connectives for prediction. For instance, the RoBERTaConn's performance will drop from 96.69\% to 55.34\% when manually-annotated connectives are not available.

\begin{figure}[t]
\centering
\includegraphics[scale=0.47,trim=0 0 0 0]{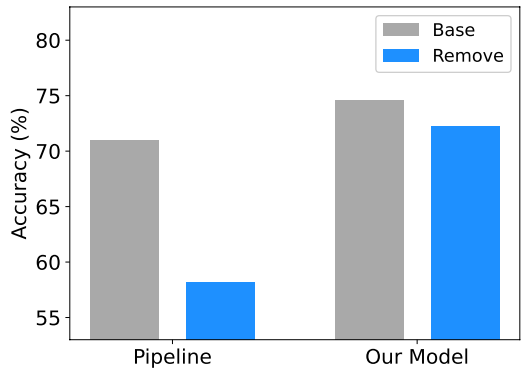}
\setlength{\abovecaptionskip}{6pt} 
\setlength{\belowcaptionskip}{0pt} 
\caption{Level1 classification results on PDTB 2.0 (Ji split). "Remove" denotes the generated connectives are removed from the original model ("Base").}
\label{figure:pdtb2_l1_acc_rm}
\end{figure}

To probe whether our model suffers such an issue, we perform the second analysis by removing the generated connectives in our model and observing changes in its performance. The same setting is applied to Pipeline for comparison. Figure \ref{figure:pdtb2_l1_acc_rm} shows the Level1 classification results\textsuperscript{\ref{foot:detail_ana}} on PDTB 2.0 (Ji split). Both models see a performance drop but still outperform RoBERTaConn. This is because these two models' relation classifiers input the generated connectives rather than the annotated ones for training, alleviating their reliance on connectives. The decrease of Our Method (74.59\% $\to$ 72.27\%) is much smaller than that of Pipeline (71.01\% $\to$ 58.15\%). We speculate that the end-to-end training enables our model to learn a good balance between arguments and discourse connectives for relation classification. By contrast, Pipeline fails to do so due to the separate training of connectives generation and relation classification.

Finally, we show in Table \ref{table:group_ana} the results of relation classifiers in Multi-Task, Pipeline, and Our method\footnote{This analysis is not performed on other models (e.g., Adversarial) because they don't generate or predict connectives.} on PDTB 2.0 when connectives are correctly and incorrectly generated or predicted. Note that these three models' results are not directly comparable in the correct and incorrect groups since their predictions on connectives are different\textsuperscript{\ref{foot:detail_ana}} (not overlap). To solve this, we calculate the perform\-ance gain of each model over the RoBERTa baseline and compare them from the gain perspective. When connectives are correctly generated, Pipeline and Our Model outperform the RoBERTa baseline by more than 10\% in accuracy, while Multi-task's improvement is only 6.9\%. This suggests that Pipeline and Our Model utilize connectives more efficiently than Multi-Task. On the other hand, when the connectives' prediction is incorrect, Pipeline's performance is worse than the RoBERTa baseline by 1.64\%. Compared to it, Multi-task and Our Method achieve comparable performance to RoBERTa, showing good robustness when exposed to incorrect connectives. Despite achieving better results than baselines in both groups, our model performs significantly worse in the incorrect connective group than in the correct one. This indicates that its major performance bottleneck originates from the incorrectly generated connectives. A possible improvement is first pre-training our model on a large explicit connectives corpus, like ~\citet{sileo-etal-2019-mining}. By doing so, the connective generation module may generate more correct connectives, thus improving classification performance, which we leave for future work.

\begin{table}[t]
\centering
\normalsize
\scalebox{0.82}{
\begin{tabular}{l|c|c}
\hline
Models     &  Correct Group \hspace{0.2em} & \hspace{0.3em} Incorrect Group \hspace{0.3em} \\ \hline
Base\textsubscript{Multi-Task}    & 83.67\hspace{3.0em}              & 59.82\hspace{2.8em}                \\
Multi-Task & 90.60(+6.93)       & 59.88(+0.06)         \\ \hline
Base\textsubscript{Pipeline}    & 78.87\hspace{3.0em}              & 61.46\hspace{2.8em}                \\
Pipeline   & 89.29(+10.4)      & 59.81(-1.64)         \\ \hline
Base\textsubscript{Our Model}    & 80.28\hspace{3.0em}              & 60.56\hspace{2.8em}                \\
Our Model & 94.04(+13.8)      & \hspace{0.15em}62.22(+1.66)         \\ \hline
\end{tabular}}
\caption{Level1 classification results on PDTB 2.0 (Ji split) when connectives are correctly and incorrectly generated (or predicted). "+" and "-" denote the increase and decrease compared to the RoBERTa baseline (Base).}
\label{table:group_ana}
\end{table}

\begin{table*}[t]
\centering
\setlength{\abovecaptionskip}{8pt} 
\setlength{\belowcaptionskip}{-8pt} 
\scalebox{0.95}{
\begin{tabular}{l|ccccc}
\hline
Labels                                    & \multicolumn{1}{l}{RoBERTa} & \multicolumn{1}{l}{Adversarial} & \multicolumn{1}{l}{Multi-Task} & \multicolumn{1}{l}{Pipeline} & \multicolumn{1}{l}{Our Model} \\ \hline
\multicolumn{1}{l|}{Temporal.Asynchronous}    & 54.62                       & 55.01                           & 58.37                          & 55.69                        & \textbf{59.48}                          \\
\multicolumn{1}{l|}{Temporal.Synchrony}       & 00.00                       & \textbf{06.03}                           & 00.00                          & 04.00                        & 00.00                          \\ \hline
\multicolumn{1}{l|}{Contingency.Cause}           & 60.03                       & 59.00                           & 64.24                          & 65.40                        & \textbf{66.35}                          \\
\multicolumn{1}{l|}{Contingency.Pragmatic cause} & 00.00                       & \textbf{05.00}                           & 00.00                          & 00.00                        & 00.00                          \\ \hline
\multicolumn{1}{l|}{Comparison.Contrast}        & 60.44                       & 58.20                           & 61.73                          & 60.78                        & \textbf{65.75}                          \\
\multicolumn{1}{l|}{Comparison.Concession}      & 00.00                       & 01.14                           & 00.00                          & \textbf{01.82}                        & 00.00                          \\ \hline
\multicolumn{1}{l|}{Expansion.Conjunction}      & 56.03                       & 53.26                           & \textbf{58.94}                          & 54.79                        & 57.04                          \\
\multicolumn{1}{l|}{Expansion.Instantiation}    & 74.07                       & 72.85                           & \textbf{74.12}                          & 70.76                        & 73.87                          \\
\multicolumn{1}{l|}{Expansion.Restatement}      & 57.87                       & 56.94                           & 59.68                          & 57.75                        & \textbf{60.94}                          \\
\multicolumn{1}{l|}{Expansion.Alternative}      & 49.06                       & 44.76                           & \textbf{54.82}                          & 43.96                        & 51.13                          \\
\multicolumn{1}{l|}{Expansion.List}             & 18.07                       & 11.68                           & 11.43                          & \textbf{29.96}                        & 25.47                          \\ \hline
\end{tabular}}
\caption{F1 results for each second-level relation of PDTB 2.0.}
\label{table:rel_ana}
\end{table*}

\subsection{Relation Analysis}
\label{sec-rel_ana}
We investigate which relations benefit from the joint training of connective generation and relation classification and compare it with other baselines. Table 4 shows different models' F1-score for each second-level sense of PDTB 2.0 (Ji split). Generally, connectives benefit the prediction of most relation types, especially in Multi-Task, Pipeline, and Our Method. For example, these three models outperform the RoBERTa baseline by more than 4\% in the F1-score on the Contingency.Cause relation. On some relations, such as Expansion.Instantiation, connective-enhanced models show different tendencies, with some experiencing improvement while others drop. Surprisingly, all models fail to predict Temporal.Synchrony, Contingency.Pragmatic cause, and Comparison.Concession despite using manually-annotated connectives during training. We speculate this is caused by their limited number of training instances, making models tend to predict other frequent labels. One feasible solution to this issue is Contrastive Learning ~\citep{chen2020simple}, which has been shown to improve the predictive performance of these three relations ~\citep{long-webber-2022-facilitating}. We leave integrating Contrastive Learning with our method to future work.

\begin{table}[t]
\centering
\scalebox{0.84}{
\begin{tabular}{l|cc|cc}
\hline
           & \multicolumn{2}{c|}{PDTB 2.0} & \multicolumn{2}{c}{PDTB 3.0} \\ \hline
Models     & \hspace{0.6em}  Acc     \hspace{0.6em}      &  \hspace{0.6em} F1   \hspace{0.6em}         & \hspace{0.6em} Acc     \hspace{0.6em}      & \hspace{0.6em} F1    \hspace{0.6em}       \\ \hline
Our Model & \textbf{74.59}         & \textbf{68.64}         & \textbf{76.23}         & \textbf{71.15}        \\ 
- SS       & 73.42              &  66.68             &   75.87            &  70.68            \\ 
- SS, $\mathcal{L}_{Conn}$       & 70.63              &  63.43             &   74.58            &  69.17            \\ \hline
RoBERTa & 68.61     & 60.89     & 73.51     & 67.98 \\ \hline
\end{tabular}}
\caption{Ablation study for \textbf{S}cheduled \textbf{S}ampling and connective generation loss $\mathcal{L}_{Conn}$.}
\label{table:ablation_study}
\end{table}

\vspace{-2pt}
\subsection{Ablation Study}
We conduct ablation studies to evaluate the effect\-iveness of Scheduled Sampling (SS) and the Connective generation loss $\mathcal{L}_{Conn}$. To this end, we test the performance of our method by first removing SS and then removing $\mathcal{L}_{Conn}$. Note that removing $\mathcal{L}_{Conn}$ means that our whole model is trained with only gradients from $\mathcal{L}_{Rel}$.

Table \ref{table:ablation_study} shows the Level1 classification results on PDTB 2.0 and PDTB 3.0 (Ji split). We can observe from the table that eliminating any of them would hurt the performance, showing their essential to achieve good performance. Surprisingly, our model training with only $\mathcal{L}_{Rel}$ performs much better than the RoBERTa baseline. This indicates that the performance gain of our full model comes not only from the training signals provided by manually-annotated connectives but also from its well-designed structure inspired by PDTB's annotation (i.e. the connective generation module and relation predict\-ion module).

\section{Conclusion}
In this paper, we propose a novel connective-enhanced method for implicit relation classification, inspired by the annotation of PDTB. We introduce several key techniques to efficiently train our model in an end-to-end manner. Experiments on three benchmarks demonstrate that our method consist\-ently outperforms various baseline models. Analyses of the models' behavior show that our approach can learn a good balance between using arguments and connectives for implicit discourse relation prediction.

\section{Limitations}
Despite achieving good performance, there are some limitations in our study. The first is how to handle ambiguous instances in the corpus. 3.45\% of the implicit data in PDTB 2.0 and 5\% in PDTB 3.0 contains more than one label. Currently, we follow previous work and simply use the first label for training. But there might be a better solution to handle those cases. Another is the required time for training. To mimic the annotation process of PDTB, our model needs to pass through the embedding layer and transformers twice, so it takes more time to train than the RoBERTa baseline. However, our training time is shorter than Pipeline and Adversarial due to those two models' pipeline setup and adversarial training strategy. Also, note that our method has a similar number of parameters to the RoBERTa baseline since we share embedding layers and transformers between the connection generation and relation classification modules in our approach. Therefore, the memory required to train our model is not much different from that required to train the RoBERTa baseline.


\bibliography{main.bbl}

\begin{thebibliography}{50}
\expandafter\ifx\csname natexlab\endcsname\relax\def\natexlab#1{#1}\fi

\bibitem[{Bengio et~al.(2015)Bengio, Vinyals, Jaitly, and
  Shazeer}]{bengio2015scheduled}
Samy Bengio, Oriol Vinyals, Navdeep Jaitly, and Noam Shazeer. 2015.
\newblock \href
  {https://proceedings.neurips.cc/paper/2015/file/e995f98d56967d946471af29d7bf99f1-Paper.pdf}
  {Scheduled sampling for sequence prediction with recurrent neural networks}.
\newblock In \emph{Advances in Neural Information Processing Systems},
  volume~28.

\bibitem[{Bourgonje(2021)}]{bourgonje:diss}
Peter Bourgonje. 2021.
\newblock \href {https://doi.org/10.25932/publishup-50663} {\emph{Shallow
  Discourse Parsing for German}}.
\newblock {D}octoral {T}hesis, {Universität Potsdam}.

\bibitem[{Bourgonje and Stede(2020)}]{bourgonje-stede-2020-potsdam}
Peter Bourgonje and Manfred Stede. 2020.
\newblock \href {https://aclanthology.org/2020.lrec-1.133} {The {P}otsdam
  commentary corpus 2.2: Extending annotations for shallow discourse parsing}.
\newblock In \emph{Proceedings of the Twelfth Language Resources and Evaluation
  Conference}, pages 1061--1066, Marseille, France. European Language Resources
  Association.

\bibitem[{Chen et~al.(2020)Chen, Kornblith, Norouzi, and
  Hinton}]{chen2020simple}
Ting Chen, Simon Kornblith, Mohammad Norouzi, and Geoffrey Hinton. 2020.
\newblock A simple framework for contrastive learning of visual
  representations.
\newblock In \emph{International {C}onference on {M}achine {L}earning}, pages
  1597--1607. PMLR.

\bibitem[{Collobert et~al.(2011)Collobert, Weston, Bottou, Karlen, Kavukcuoglu,
  and Kuksa}]{collobert2011natural}
Ronan Collobert, Jason Weston, L\'{e}on Bottou, Michael Karlen, Koray
  Kavukcuoglu, and Pavel Kuksa. 2011.
\newblock \href {https://dl.acm.org/doi/10.5555/1953048.2078186} {Natural
  language processing (almost) from scratch}.
\newblock \emph{Journal of {M}achine {L}earning {R}esearch}, 12:2493–2537.

\bibitem[{Devlin et~al.(2019)Devlin, Chang, Lee, and
  Toutanova}]{devlin-etal-2019-bert}
Jacob Devlin, Ming-Wei Chang, Kenton Lee, and Kristina Toutanova. 2019.
\newblock \href {https://doi.org/10.18653/v1/N19-1423} {{BERT}: Pre-training of
  deep bidirectional transformers for language understanding}.
\newblock In \emph{Proceedings of the 2019 Conference of the North {A}merican
  Chapter of the Association for Computational Linguistics: Human Language
  Technologies, Volume 1 (Long and Short Papers)}, pages 4171--4186,
  Minneapolis, Minnesota. Association for Computational Linguistics.

\bibitem[{Guo et~al.(2018)Guo, He, Jin, Dang, Wang, and
  Li}]{guo-etal-2018-implicit}
Fengyu Guo, Ruifang He, Di~Jin, Jianwu Dang, Longbiao Wang, and Xiangang Li.
  2018.
\newblock \href {https://aclanthology.org/C18-1046} {Implicit discourse
  relation recognition using neural tensor network with interactive attention
  and sparse learning}.
\newblock In \emph{Proceedings of the 27th International Conference on
  Computational Linguistics}, pages 547--558, Santa Fe, New Mexico, USA.
  Association for Computational Linguistics.

\bibitem[{Habernal and Gurevych(2017)}]{habernal-gurevych-2017-argumentation}
Ivan Habernal and Iryna Gurevych. 2017.
\newblock \href {https://doi.org/10.1162/COLI_a_00276} {Argumentation mining in
  user-generated web discourse}.
\newblock \emph{Computational Linguistics}, 43(1):125--179.

\bibitem[{Hewett et~al.(2019)Hewett, Prakash~Rane, Harlacher, and
  Stede}]{hewett-etal-2019-utility}
Freya Hewett, Roshan Prakash~Rane, Nina Harlacher, and Manfred Stede. 2019.
\newblock \href {https://doi.org/10.18653/v1/W19-4512} {The utility of
  discourse parsing features for predicting argumentation structure}.
\newblock In \emph{Proceedings of the 6th Workshop on Argument Mining}, pages
  98--103, Florence, Italy. Association for Computational Linguistics.

\bibitem[{Jang et~al.(2017)Jang, Gu, and Poole}]{gumbel}
Eric Jang, Shixiang Gu, and Ben Poole. 2017.
\newblock \href {https://openreview.net/forum?id=rkE3y85ee} {Categorical
  reparameterization with gumbel-softmax}.
\newblock In \emph{5th International Conference on Learning Representations,
  {ICLR} 2017}.

\bibitem[{Ji and Eisenstein(2015)}]{ji-eisenstein-2015-one}
Yangfeng Ji and Jacob Eisenstein. 2015.
\newblock \href {https://doi.org/10.1162/tacl_a_00142} {One vector is not
  enough: Entity-augmented distributed semantics for discourse relations}.
\newblock \emph{Transactions of the Association for Computational Linguistics},
  3:329--344.

\bibitem[{Jiang et~al.(2021)Jiang, Qian, Chen, Tang, Zhan, and
  Zhan}]{pipejiang}
Congcong Jiang, Tieyun Qian, Zhuang Chen, Kejian Tang, Shaohui Zhan, and Tao
  Zhan. 2021.
\newblock Generating pseudo connectives with mlms for implicit discourse
  relation recognition.
\newblock In \emph{PRICAI 2021: Trends in Artificial Intelligence}, pages
  113--126, Cham. Springer International Publishing.

\bibitem[{Kim et~al.(2020)Kim, Feng, Gunasekara, and
  Lastras}]{kim-etal-2020-implicit}
Najoung Kim, Song Feng, Chulaka Gunasekara, and Luis Lastras. 2020.
\newblock \href {https://doi.org/10.18653/v1/2020.acl-main.480} {Implicit
  discourse relation classification: We need to talk about evaluation}.
\newblock In \emph{Proceedings of the 58th Annual Meeting of the Association
  for Computational Linguistics}, pages 5404--5414, Online. Association for
  Computational Linguistics.

\bibitem[{Kishimoto et~al.(2018)Kishimoto, Murawaki, and
  Kurohashi}]{kishimoto-etal-2018-knowledge}
Yudai Kishimoto, Yugo Murawaki, and Sadao Kurohashi. 2018.
\newblock \href {https://aclanthology.org/C18-1049} {A knowledge-augmented
  neural network model for implicit discourse relation classification}.
\newblock In \emph{Proceedings of the 27th International Conference on
  Computational Linguistics}, pages 584--595, Santa Fe, New Mexico, USA.
  Association for Computational Linguistics.

\bibitem[{Kishimoto et~al.(2020)Kishimoto, Murawaki, and
  Kurohashi}]{kishimoto-etal-2020-adapting}
Yudai Kishimoto, Yugo Murawaki, and Sadao Kurohashi. 2020.
\newblock \href {https://aclanthology.org/2020.lrec-1.145} {Adapting {BERT} to
  implicit discourse relation classification with a focus on discourse
  connectives}.
\newblock In \emph{Proceedings of the Twelfth Language Resources and Evaluation
  Conference}, pages 1152--1158, Marseille, France. European Language Resources
  Association.

\bibitem[{Kurfal{\i} and {\"O}stling(2021)}]{kurfali-ostling-2021-lets}
Murathan Kurfal{\i} and Robert {\"O}stling. 2021.
\newblock \href {https://doi.org/10.18653/v1/2021.unimplicit-1.1} {Let{'}s be
  explicit about that: Distant supervision for implicit discourse relation
  classification via connective prediction}.
\newblock In \emph{Proceedings of the 1st Workshop on Understanding Implicit
  and Underspecified Language}, pages 1--10, Online. Association for
  Computational Linguistics.

\bibitem[{Lei et~al.(2017)Lei, Wang, Liu, Ilievski, He, and Kan}]{ijcai-lei}
Wenqiang Lei, Xuancong Wang, Meichun Liu, Ilija Ilievski, Xiangnan He, and
  Min-Yen Kan. 2017.
\newblock \href {https://doi.org/10.24963/ijcai.2017/562} {Swim: A simple word
  interaction model for implicit discourse relation recognition}.
\newblock In \emph{Proceedings of the Twenty-Sixth International Joint
  Conference on Artificial Intelligence, {IJCAI-17}}, pages 4026--4032.

\bibitem[{Lin et~al.(2009)Lin, Kan, and Ng}]{lin-etal-2009-recognizing}
Ziheng Lin, Min-Yen Kan, and Hwee~Tou Ng. 2009.
\newblock \href {https://aclanthology.org/D09-1036} {Recognizing implicit
  discourse relations in the {P}enn {D}iscourse {T}reebank}.
\newblock In \emph{Proceedings of the 2009 Conference on Empirical Methods in
  Natural Language Processing}, pages 343--351, Singapore. Association for
  Computational Linguistics.

\bibitem[{Lin et~al.(2011)Lin, Ng, and Kan}]{lin-etal-2011-automatically}
Ziheng Lin, Hwee~Tou Ng, and Min-Yen Kan. 2011.
\newblock \href {https://aclanthology.org/P11-1100} {Automatically evaluating
  text coherence using discourse relations}.
\newblock In \emph{Proceedings of the 49th Annual Meeting of the Association
  for Computational Linguistics: Human Language Technologies}, pages 997--1006,
  Portland, Oregon, USA. Association for Computational Linguistics.

\bibitem[{Lin et~al.(2014)Lin, Ng, and Kan}]{lin2014pdtb}
Ziheng Lin, Hwee~Tou Ng, and Min-Yen Kan. 2014.
\newblock A {PDTB}-styled end-to-end discourse parser.
\newblock \emph{Natural Language Engineering}, 20(2):151--184.

\bibitem[{Liu et~al.(2020)Liu, Ou, Song, and Jiang}]{BMGF}
Xin Liu, Jiefu Ou, Yangqiu Song, and Xin Jiang. 2020.
\newblock \href {https://doi.org/10.24963/ijcai.2020/530} {On the importance of
  word and sentence representation learning in implicit discourse relation
  classification}.
\newblock In \emph{Proceedings of the Twenty-Ninth International Joint
  Conference on Artificial Intelligence, {IJCAI-20}}, pages 3830--3836.
  International Joint Conferences on Artificial Intelligence Organization.
\newblock Main track.

\bibitem[{Liu and Li(2016)}]{liu-li-2016-recognizing}
Yang Liu and Sujian Li. 2016.
\newblock \href {https://doi.org/10.18653/v1/D16-1130} {Recognizing implicit
  discourse relations via repeated reading: Neural networks with multi-level
  attention}.
\newblock In \emph{Proceedings of the 2016 Conference on Empirical Methods in
  Natural Language Processing}, pages 1224--1233, Austin, Texas. Association
  for Computational Linguistics.

\bibitem[{Liu et~al.(2019)Liu, Ott, Goyal, Du, Joshi, Chen, Levy, Lewis,
  Zettlemoyer, and Stoyanov}]{roberta}
Yinhan Liu, Myle Ott, Naman Goyal, Jingfei Du, Mandar Joshi, Danqi Chen, Omer
  Levy, Mike Lewis, Luke Zettlemoyer, and Veselin Stoyanov. 2019.
\newblock \href {http://arxiv.org/abs/1907.11692} {{RoBERTa}: {A} robustly
  optimized {BERT} {P}retraining {A}pproach}.
\newblock \emph{CoRR}, abs/1907.11692.

\bibitem[{Long and Webber(2022)}]{long-webber-2022-facilitating}
Wanqiu Long and Bonnie Webber. 2022.
\newblock \href {https://aclanthology.org/2022.emnlp-main.734} {Facilitating
  contrastive learning of discourse relational senses by exploiting the
  hierarchy of sense relations}.
\newblock In \emph{Proceedings of the 2022 Conference on Empirical Methods in
  Natural Language Processing}, pages 10704--10716, Abu Dhabi, United Arab
  Emirates. Association for Computational Linguistics.

\bibitem[{Longyue(2019)}]{longyue2019discourse}
Wang Longyue. 2019.
\newblock \emph{Discourse-aware neural machine translation}.
\newblock Ph.D. thesis, Dublin City University.

\bibitem[{Meyer(2015)}]{meyer2015discourse}
Thomas Meyer. 2015.
\newblock \emph{Discourse-level features for statistical machine translation}.
\newblock Ph.D. thesis, École polytechnique fédérale de Lausanne (EPFL).

\bibitem[{Mihaylov and Frank(2019)}]{mihaylov-frank-2019-discourse}
Todor Mihaylov and Anette Frank. 2019.
\newblock \href {https://doi.org/10.18653/v1/D19-1257} {Discourse-aware
  semantic self-attention for narrative reading comprehension}.
\newblock In \emph{Proceedings of the 2019 Conference on Empirical Methods in
  Natural Language Processing and the 9th International Joint Conference on
  Natural Language Processing (EMNLP-IJCNLP)}, pages 2541--2552, Hong Kong,
  China. Association for Computational Linguistics.

\bibitem[{Miwa and Bansal(2016)}]{miwa-bansal-2016-end}
Makoto Miwa and Mohit Bansal. 2016.
\newblock \href {https://doi.org/10.18653/v1/P16-1105} {End-to-end relation
  extraction using {LSTM}s on sequences and tree structures}.
\newblock In \emph{Proceedings of the 54th Annual Meeting of the Association
  for Computational Linguistics (Volume 1: Long Papers)}, pages 1105--1116,
  Berlin, Germany. Association for Computational Linguistics.

\bibitem[{Pitler et~al.(2009)Pitler, Louis, and
  Nenkova}]{pitler-etal-2009-automatic}
Emily Pitler, Annie Louis, and Ani Nenkova. 2009.
\newblock \href {https://aclanthology.org/P09-1077} {Automatic sense prediction
  for implicit discourse relations in text}.
\newblock In \emph{Proceedings of the Joint Conference of the 47th Annual
  Meeting of the {ACL} and the 4th International Joint Conference on Natural
  Language Processing of the {AFNLP}}, pages 683--691, Suntec, Singapore.
  Association for Computational Linguistics.

\bibitem[{Pitler and Nenkova(2009)}]{pitler-nenkova-2009-using}
Emily Pitler and Ani Nenkova. 2009.
\newblock \href {https://aclanthology.org/P09-2004} {Using syntax to
  disambiguate explicit discourse connectives in text}.
\newblock In \emph{Proceedings of the {ACL}-{IJCNLP} 2009 Conference Short
  Papers}, pages 13--16, Suntec, Singapore. Association for Computational
  Linguistics.

\bibitem[{Prasad et~al.(2008)Prasad, Dinesh, Lee, Miltsakaki, Robaldo, Joshi,
  and Webber}]{prasad-etal-2008-penn}
Rashmi Prasad, Nikhil Dinesh, Alan Lee, Eleni Miltsakaki, Livio Robaldo,
  Aravind Joshi, and Bonnie Webber. 2008.
\newblock \href
  {http://www.lrec-conf.org/proceedings/lrec2008/pdf/754_paper.pdf} {The {P}enn
  {D}iscourse {T}ree{B}ank 2.0.}
\newblock In \emph{Proceedings of the Sixth International Conference on
  Language Resources and Evaluation ({LREC}'08)}, Marrakech, Morocco. European
  Language Resources Association (ELRA).

\bibitem[{Qin et~al.(2017)Qin, Zhang, Zhao, Hu, and
  Xing}]{qin-etal-2017-adversarial}
Lianhui Qin, Zhisong Zhang, Hai Zhao, Zhiting Hu, and Eric Xing. 2017.
\newblock \href {https://doi.org/10.18653/v1/P17-1093} {Adversarial
  connective-exploiting networks for implicit discourse relation
  classification}.
\newblock In \emph{Proceedings of the 55th Annual Meeting of the Association
  for Computational Linguistics (Volume 1: Long Papers)}, pages 1006--1017,
  Vancouver, Canada. Association for Computational Linguistics.

\bibitem[{Ranzato et~al.(2016)Ranzato, Chopra, Auli, and
  Zaremba}]{RanzatoCAZ15}
Marc'Aurelio Ranzato, Sumit Chopra, Michael Auli, and Wojciech Zaremba. 2016.
\newblock \href {http://arxiv.org/abs/1511.06732} {Sequence level training with
  recurrent neural networks}.
\newblock In \emph{4th International Conference on Learning Representations,
  {ICLR} 2016}.

\bibitem[{Rutherford and Xue(2014)}]{rutherford-xue-2014-discovering}
Attapol Rutherford and Nianwen Xue. 2014.
\newblock \href {https://doi.org/10.3115/v1/E14-1068} {Discovering implicit
  discourse relations through brown cluster pair representation and coreference
  patterns}.
\newblock In \emph{Proceedings of the 14th Conference of the {E}uropean Chapter
  of the Association for Computational Linguistics}, pages 645--654,
  Gothenburg, Sweden. Association for Computational Linguistics.

\bibitem[{Shi et~al.(2021)Shi, Ding, Du, Liu, and Qin}]{shi-etal-2021-neural}
Jihao Shi, Xiao Ding, Li~Du, Ting Liu, and Bing Qin. 2021.
\newblock \href {https://doi.org/10.18653/v1/2021.emnlp-main.298} {Neural
  natural logic inference for interpretable question answering}.
\newblock In \emph{Proceedings of the 2021 Conference on Empirical Methods in
  Natural Language Processing}, pages 3673--3684, Online and Punta Cana,
  Dominican Republic. Association for Computational Linguistics.

\bibitem[{Shi and Demberg(2019)}]{shi-demberg-2019-learning}
Wei Shi and Vera Demberg. 2019.
\newblock \href {https://doi.org/10.18653/v1/W19-0416} {Learning to explicitate
  connectives with {S}eq2{S}eq network for implicit discourse relation
  classification}.
\newblock In \emph{Proceedings of the 13th International Conference on
  Computational Semantics - Long Papers}, pages 188--199, Gothenburg, Sweden.
  Association for Computational Linguistics.

\bibitem[{Shi et~al.(2017)Shi, Yung, Rubino, and Demberg}]{shi-etal-2017-using}
Wei Shi, Frances Yung, Raphael Rubino, and Vera Demberg. 2017.
\newblock \href {https://aclanthology.org/I17-1049} {Using explicit discourse
  connectives in translation for implicit discourse relation classification}.
\newblock In \emph{Proceedings of the Eighth International Joint Conference on
  Natural Language Processing (Volume 1: Long Papers)}, pages 484--495, Taipei,
  Taiwan. Asian Federation of Natural Language Processing.

\bibitem[{Sileo et~al.(2019)Sileo, Van De~Cruys, Pradel, and
  Muller}]{sileo-etal-2019-mining}
Damien Sileo, Tim Van De~Cruys, Camille Pradel, and Philippe Muller. 2019.
\newblock \href {https://doi.org/10.18653/v1/N19-1351} {Mining discourse
  markers for unsupervised sentence representation learning}.
\newblock In \emph{Proceedings of the 2019 Conference of the North {A}merican
  Chapter of the Association for Computational Linguistics: Human Language
  Technologies, Volume 1 (Long and Short Papers)}, pages 3477--3486,
  Minneapolis, Minnesota. Association for Computational Linguistics.

\bibitem[{S{\o}gaard and Goldberg(2016)}]{sogaard-goldberg-2016-deep}
Anders S{\o}gaard and Yoav Goldberg. 2016.
\newblock \href {https://doi.org/10.18653/v1/P16-2038} {Deep multi-task
  learning with low level tasks supervised at lower layers}.
\newblock In \emph{Proceedings of the 54th Annual Meeting of the Association
  for Computational Linguistics (Volume 2: Short Papers)}, pages 231--235,
  Berlin, Germany. Association for Computational Linguistics.

\bibitem[{Sporleder and Lascarides(2008)}]{sporleder_lascarides_2008}
Caroline Sporleder and Alex Lascarides. 2008.
\newblock \href {https://doi.org/10.1017/S1351324906004451} {Using
  automatically labelled examples to classify rhetorical relations: an
  assessment}.
\newblock \emph{Natural Language Engineering}, 14(3):369–416.

\bibitem[{Strubell et~al.(2018)Strubell, Verga, Andor, Weiss, and
  McCallum}]{strubell-etal-2018-linguistically}
Emma Strubell, Patrick Verga, Daniel Andor, David Weiss, and Andrew McCallum.
  2018.
\newblock \href {https://doi.org/10.18653/v1/D18-1548} {Linguistically-informed
  self-attention for semantic role labeling}.
\newblock In \emph{Proceedings of the 2018 Conference on Empirical Methods in
  Natural Language Processing}, pages 5027--5038, Brussels, Belgium.
  Association for Computational Linguistics.

\bibitem[{Webber et~al.(2019)Webber, Prasad, Lee, and Joshi}]{webber2019penn}
Bonnie Webber, Rashmi Prasad, Alan Lee, and Aravind Joshi. 2019.
\newblock The {P}enn {D}iscourse {T}ree{B}ank 3.0 annotation manual.
\newblock \emph{Philadelphia, University of Pennsylvania}, 35:108.

\bibitem[{Wu et~al.(2022)Wu, Cao, Ge, Liu, Zhang, and Su}]{LDSGM}
Changxing Wu, Liuwen Cao, Yubin Ge, Yang Liu, Min Zhang, and Jinsong Su. 2022.
\newblock \href {https://doi.org/10.1609/aaai.v36i10.21401} {A label
  dependence-aware sequence generation model for multi-level implicit discourse
  relation recognition}.
\newblock \emph{Proceedings of the AAAI Conference on Artificial Intelligence},
  36(10):11486--11494.

\bibitem[{Xiang and Wang(2023)}]{xiang2022survey}
Wei Xiang and Bang Wang. 2023.
\newblock \href {https://doi.org/10.1145/3574134} {A survey of implicit
  discourse relation recognition}.
\newblock \emph{ACM Computing Surveys}, 55(12):1--34.

\bibitem[{Xiang et~al.(2022)Xiang, Wang, Dai, and
  Wang}]{xiang-etal-2022-connprompt}
Wei Xiang, Zhenglin Wang, Lu~Dai, and Bang Wang. 2022.
\newblock \href {https://aclanthology.org/2022.coling-1.75} {{C}onn{P}rompt:
  Connective-cloze prompt learning for implicit discourse relation
  recognition}.
\newblock In \emph{Proceedings of the 29th International Conference on
  Computational Linguistics}, pages 902--911, Gyeongju, Republic of Korea.
  International Committee on Computational Linguistics.

\bibitem[{Xue et~al.(2016)Xue, Ng, Pradhan, Rutherford, Webber, Wang, and
  Wang}]{xue-etal-2016-conll}
Nianwen Xue, Hwee~Tou Ng, Sameer Pradhan, Attapol Rutherford, Bonnie Webber,
  Chuan Wang, and Hongmin Wang. 2016.
\newblock \href {https://doi.org/10.18653/v1/K16-2001} {{C}o{NLL} 2016 shared
  task on multilingual shallow discourse parsing}.
\newblock In \emph{Proceedings of the {C}o{NLL}-16 shared task}, pages 1--19,
  Berlin, Germany. Association for Computational Linguistics.

\bibitem[{Zhang et~al.(2015)Zhang, Su, Xiong, Lu, Duan, and
  Yao}]{zhang-etal-2015-shallow}
Biao Zhang, Jinsong Su, Deyi Xiong, Yaojie Lu, Hong Duan, and Junfeng Yao.
  2015.
\newblock \href {https://doi.org/10.18653/v1/D15-1266} {Shallow convolutional
  neural network for implicit discourse relation recognition}.
\newblock In \emph{Proceedings of the 2015 Conference on Empirical Methods in
  Natural Language Processing}, pages 2230--2235, Lisbon, Portugal. Association
  for Computational Linguistics.

\bibitem[{Zhang(2018)}]{zhang-2018-joint}
Yue Zhang. 2018.
\newblock \href {https://aclanthology.org/D18-3001} {Joint models for {NLP}}.
\newblock In \emph{Proceedings of the 2018 Conference on Empirical Methods in
  Natural Language Processing: Tutorial Abstracts}, Melbourne, Australia.
  Association for Computational Linguistics.

\bibitem[{Zhou et~al.(2022)Zhou, Lan, Wu, Chen, and
  Ma}]{zhou-etal-2022-prompt-based}
Hao Zhou, Man Lan, Yuanbin Wu, Yuefeng Chen, and Meirong Ma. 2022.
\newblock \href {https://aclanthology.org/2022.findings-emnlp.282}
  {Prompt-based connective prediction method for fine-grained implicit
  discourse relation recognition}.
\newblock In \emph{Findings of the Association for Computational Linguistics:
  EMNLP 2022}, pages 3848--3858, Abu Dhabi, United Arab Emirates. Association
  for Computational Linguistics.

\bibitem[{Zhou et~al.(2010)Zhou, Lan, Niu, Xu, and Su}]{zhou-etal-2010-effects}
Zhi~Min Zhou, Man Lan, Zheng~Yu Niu, Yu~Xu, and Jian Su. 2010.
\newblock \href {https://aclanthology.org/W10-4326} {The effects of discourse
  connectives prediction on implicit discourse relation recognition}.
\newblock In \emph{Proceedings of the {SIGDIAL} 2010 Conference}, pages
  139--146, Tokyo, Japan. Association for Computational Linguistics.

\end{thebibliography}
\bibliographystyle{acl_natbib}

\clearpage

\appendix
\section{Data Description}
\label{sec:data}

\begin{table}[]
\centering
\scalebox{0.72}{
\begin{tabular}{lll} \hline
                     & PDTB 2.0                    & PDTB 3.0                  \\ \hline
\multirow{4}{*}{L1}  & Comparison                  & Comparison                \\
                     & Contingency                 & Contingency               \\
                     & Expansion                   & Expansion                 \\
                     & Temporal                    & Temporal                  \\ \hline
\multirow{14}{*}{L2} & Comparison.Concession       & Comparison.Concession     \\
                     & Comparison.Contrast         & Comparison.Contrast       \\
                     & Contingency.Cause           & Contingency.Cause         \\
                     & Contingency.Pragmatic cause & Contingency.Cause+Belief  \\
                     & Expansion.Conjunction       & Contingency.Condition     \\
                     & Expansion.Instantiation     & Contingency.Purpose       \\
                     & Expansion.Alternative       & Expansion.Conjunction     \\
                     & Expansion.List              & Expansion.Equivalence     \\
                     & Expansion.Restatement       & Expansion.Instantiation   \\
                     & Temporal.Asynchronous       & Expansion.Level-of-detail \\
                     & Temporal.Synchrony          & Expansion.Manner          \\
                     &                             & Expansion.Substitution    \\
                     &                             & Temporal.Asynchronous     \\
                     &                             & Temporal.Synchronous   \\  \hline
\end{tabular}}
\setlength{\belowcaptionskip}{-5pt}
\caption{Top-level (L1) and second-level (L2) relations of PDTB 2.0 and PDTB 3.0 used in our experiments.}
\label{table:relations}
\end{table}

\begin{table}[t]
\centering
\scalebox{0.92}{
\begin{tabular}{l|ccc} \hline
         & \hspace{0.9em}Train\hspace{1.0em} & \hspace{1.0em}Dev\hspace{1.0em}  & \hspace{1.2em}Test\hspace{1.3em} \\ \hline
PDTB 2.0 & 12632 & 1183 & 1046 \\ 
PDTB 3.0 & 17085 & 1653 & 1474 \\ \hline
\end{tabular}}
\caption{Dataset statistics for the "Ji" split.}
\label{table:ji_stat}
\end{table}

The Penn Discourse TreeBank (PDTB) is the most common corpus for the task of implicit discourse relation classification. The annotation of this corpus follows a specific strategy, which consists of inserting a connective that best conveys the inferred relation, and annotating the relation label based on both the inserted implicit connectives and contextual semantics. ~\citet{prasad-etal-2008-penn} claimed that this annotation strategy significantly improves the inter-annotator agreement. PDTB has two widely used versions, PDTB 2.0 ~\citep{prasad-etal-2008-penn} and PDTB 3.0 ~\citep{webber2019penn}. In both versions, instances are annotated with senses\footnote{Some instances in PDTB have more than one label. We follow previous work to use the first label for training. While evaluating, a prediction is regarded as correct if it matches one of the annotated labels ~\citep{xue-etal-2016-conll}.} from a three-level sense hierarchy. We follow previous work ~\citep{ji-eisenstein-2015-one,kim-etal-2020-implicit} to use top-level 4-way and second-level 11-way classification for PDTB 2.0, and top-level 4-way and second-level 14-way for PDTB 3.0, and show these relations in Table \ref{table:relations}. We show the statistics information of ~\citet{ji-eisenstein-2015-one} and ~\citet{kim-etal-2020-implicit} in Tables \ref{table:ji_stat} and \ref{table:xval_stat}, respectively.

\begin{table}[t]
\centering
\scalebox{0.66}{
\begin{tabular}{llcc} \hline
fold & splitting             & PDTB 2.0            & PDTB 3.0            \\ \hline
1    & 0-1 / 2-22 / 23-24    & 1183 / 13678 / 1192 & 1653 / 18559 / 1615 \\
2    & 2-3 / 4-24 / 0-1      & 1154 / 13716 / 1183 & 1579 / 18595 / 1653 \\
3    & 4-5 / 6-1 / 2-3       & 1527 / 13372 / 1154 & 2039 / 18209 / 1579 \\
4    & 6-7 / 8-3 / 4-5       & 1247 / 13279 / 1527 & 1730 / 18058 / 2039 \\
5    & 8-9 / 10-5 / 6 -7     & 881 / 13925 / 1247  & 1138 / 18959 / 1730 \\
6    & 10-11 / 12-7 / 8-9    & 1452 / 13720 / 881  & 1944 / 18745 / 1138 \\
7    & 12-13 / 14-9 / 10-11  & 1589 / 13012 / 1452 & 2203 / 17680 / 1944 \\
8    & 14-15 / 16-11 / 12-13 & 1434 / 13030 / 1589 & 1940 / 17684 / 2203 \\
9    & 16-17 / 18-13 / 14-15 & 1480 / 13139 / 1434 & 2011 / 17876 / 1940 \\
10   & 18-19 / 20-15 / 16-17 & 1241 / 13332 / 1480 & 1667 / 18149 / 2011 \\
11   & 20-21 / 22-17 / 18-19 & 1151 / 13661 / 1241 & 1585 / 18575 / 1667 \\
12   & 22-23 / 24-19 / 20-21 & 1291 / 13611 / 1151 & 1733 / 18509 / 1585 \\ \hline
\end{tabular}}
\setlength{\belowcaptionskip}{-5pt}
\caption{Dataset statistics in cross-validation (Xval). Numbers are arranged in Dev/Train/Test order. Sections 6-1 denote sections 6-24 and sections 0-1.}
\label{table:xval_stat}
\end{table}

\begin{table}[t]
\centering
\scalebox{0.88}{
\begin{tabular}{ll}
\hline
Comparison.Concession     & Comparison.Contrast     \\
Contingency.Cause         & Expansion.Conjunction   \\
Expansion.Equivalence     & Expansion.Instantiation \\
Expansion.Level-of-detail & Temporal.Asynchronous   \\ \hline
\end{tabular}}
\setlength{\belowcaptionskip}{-5pt}
\caption{Second-level (L2) relations of PCC used in our experiments.}
\label{table:pcc_rel}
\end{table}

The Potsdam Commentary Corpus (PCC) is a German corpus constructed following the annotation guideline of PDTB ~\citep{bourgonje-stede-2020-potsdam}. In this dataset, relations are also organized in a three-level hierarchy structure. However, this corpus is relatively small, containing only 905 implicit data, and the distribution of its relations is highly uneven, especially the top-level relations. For example, the "Expansion" (540) and "Contingency" (246) account for more than 86\% of the data among all top-level relations. ~\citet{bourgonje:diss} concluded that two of four relations were never predicted in his classifier due to the highly uneven distribution of the top-level relation data. Therefore, we only use the second-level relations in our experiments. Furthermore, we use a similar setup to PDTBs for PCC, considering only relations whose frequency is not too low (over 10 in our setting). The final PCC used in our experiments contains 891 data covering 8 relations (shown in Table \ref{table:pcc_rel}). As for connectives, here, we only consider connectives with a frequency of at least 5 due to the limited size of this corpus.  

\begin{table}[t]
\centering
\scalebox{0.74}{
\begin{tabular}{llll} \hline
Hyperparam          & Value  & Hyperparam   & Value \\ \hline
Learning Rate       & 1e-5   & Batch Size   & 16    \\
Weigh Decay         & 0.1    & Max Epochs   & 10    \\
LR Decay & Linear & Warmup Ratio & 0.06  \\
Gradient Clipping   & 2.0    & Max Seq Length & 256              \\ 
$\tau$ in Equation (\ref{equ:gumbel}) & 1.0
 &  $k$ in Equation (\ref{equ:scheduled}) & 100, 200, 10  \\ \hline
\end{tabular}}
\caption{Hyperparameters for training our model.}
\label{table:hyper}
\end{table}

\begin{table}[t]
\centering
\scalebox{0.87}{
\setlength{\tabcolsep}{2pt}
\begin{tabular}{l|cc}
\hline
            & \multicolumn{2}{c}{PDTB 2.0} \\ \hline
Models      & \hspace{2.4em}Acc   \hspace{2.4em}        & \hspace{2.4em} F1    \hspace{2.4em}        \\ \hline
RoBERTaConn \hspace{1.2em} & 96.69(+41.3)         & 95.58(+58.1)         \\
Adversarial & 74.93(+5.50)         & 68.62(+6.18)         \\
Multi-Task  & 76.53(+5.71)         & 70.65(+6.86)         \\
Pipeline    & 90.13(+19.1)         & 89.13(+24.5)         \\ \hline
Our Model  & 83.71(+9.12)         & 79.25(+10.6)         \\ \hline
\end{tabular}}
\caption{Level1 classification results on PDTB 2.0 (Ji split) when manually-annotated connectives are fed to connectives enhanced models. The numbers in brackets are performance gains compared to the default settings.}
\label{table:gt_conn}
\end{table}

\begin{table}[t!]
\centering
\scalebox{0.90}{
\setlength{\tabcolsep}{3pt}
\begin{tabular}{l|cc}
\hline
           & \multicolumn{2}{c}{PDTB 2.0} \\ \hline
Models     & \hspace{2.4em}Acc    \hspace{2.4em}       & \hspace{2.4em} F1    \hspace{2.4em}        \\ \hline
Pipeline   & 58.15(-12.9)  & 46.68(-17.9)  \\ \hline
Our Model \hspace{1.2em} & 72.27(-2.32)  & 65.49(-3.15)  \\ \hline
\end{tabular}}
\caption{Level1 classification results on PDTB 2.0 (Ji split) when generated connectives are removed from Pipeline and Our Method. The numbers in brackets are performance drops compared to the default settings.}
\label{table:remove}
\end{table}

\section{Implementation Details}
\label{sec:setting}
Table \ref{table:hyper} shows the hyperparameter values for our model, most of which follow the default settings of RoBERTa ~\citep{roberta}. The value of temperature $\tau$ adopts from the default setting in Gumbel-Softmax. The $k$ in inverse sigmoid decay is set to 100 for PDTB 2.0, 200 for PDTB 3.0, and 10 for PCC. We use different $k$ for the three datasets because of their different sizes, and bigger datasets are assigned larger values. For a fair comparison, we equip baseline models with the same $\rm RoBERTa_{base}$\footnote{English version of RoBERTa-Base: \url{https://huggingface.co/roberta-base}}\footnote{German version: \url{https://huggingface.co/benjamin/roberta-base-wechsel-german}} as our method and apply the same experimental settings (e.g. GPU, optimizer, learning rate, batch size, etc.) to them. For baselines that contain model-specific hyperparameters, such as the adversarial model ~\citep{qin-etal-2017-adversarial}, we follow their default setting described in the paper. 

Considering the variability of training on PDTB, we report the mean performance of 5 random restarts for the "Ji" split ~\citep{ji-eisenstein-2015-one} and that of section-level cross-validation (Xval) like ~\citet{kim-etal-2020-implicit}. For PCC, we perform a 5-fold cross-validation on this corpus due to its limited number of data and report the mean results. We conduct all experiments on a single Tesla P40 GPU with 24GB memory. It takes about 110 minutes to train our model on every fold of PDTB 2.0, 150 minutes on every fold of PDTB 3.0, and 5 minutes on every fold of PCC.

For evaluation, we follow previous work ~\citep{ji-eisenstein-2015-one} to use accuracy (Acc, \%) and F1-macro (F1, \%) as metrics in our experiments. 

\begin{figure}[t]
\centering
\includegraphics[scale=0.48,trim=0 0 0 0]{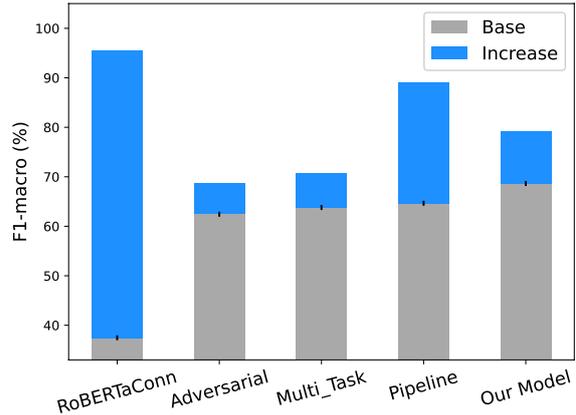}
\setlength{\abovecaptionskip}{5pt} 
\setlength{\belowcaptionskip}{0pt} 
\caption{Level1 classification results (F1) on PDTB 2.0 (Ji split) when annotated connectives are fed to models. "Increase" denotes performance gain compared to the model with default settings ("Base").}
\label{figure:pdtb2_l1_f1_inc}
\end{figure}

\begin{figure}[t]
\centering
\includegraphics[scale=0.42,trim=0 0 0 0]{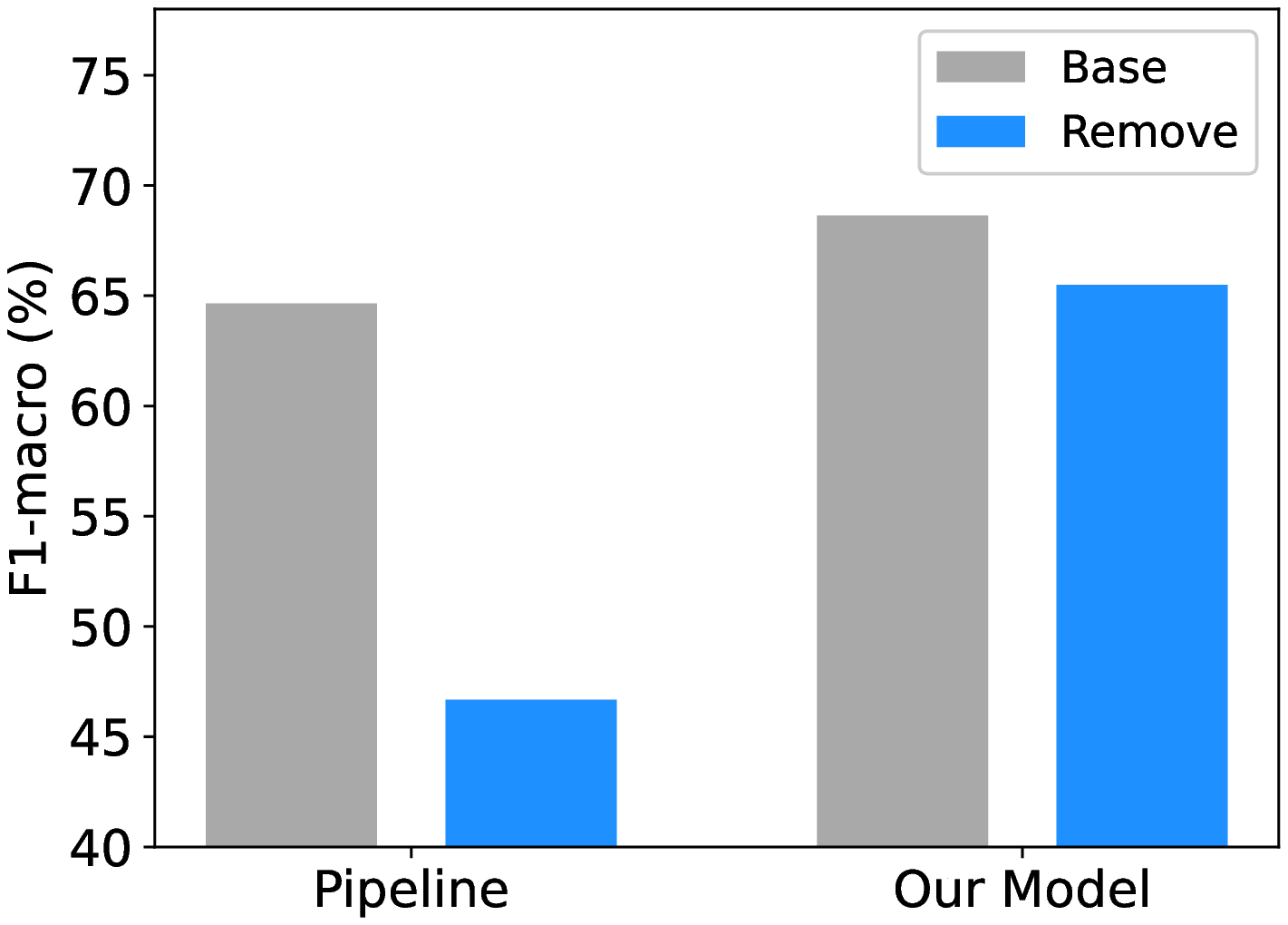}
\setlength{\abovecaptionskip}{5pt} 
\setlength{\belowcaptionskip}{0pt} 
\caption{Level1 classification results (F1) on PDTB 2.0 (Ji split). "Remove" denotes the generated connectives are removed from the original model ("Base").}
\label{figure:pdtb2_l1_f1_rm}
\end{figure}

\section{Performance Analysis}
\label{app:sec:detailed_behavior}

\begin{figure*}[t!]
\centering
\includegraphics[scale=0.53,trim=0 0 0 0]{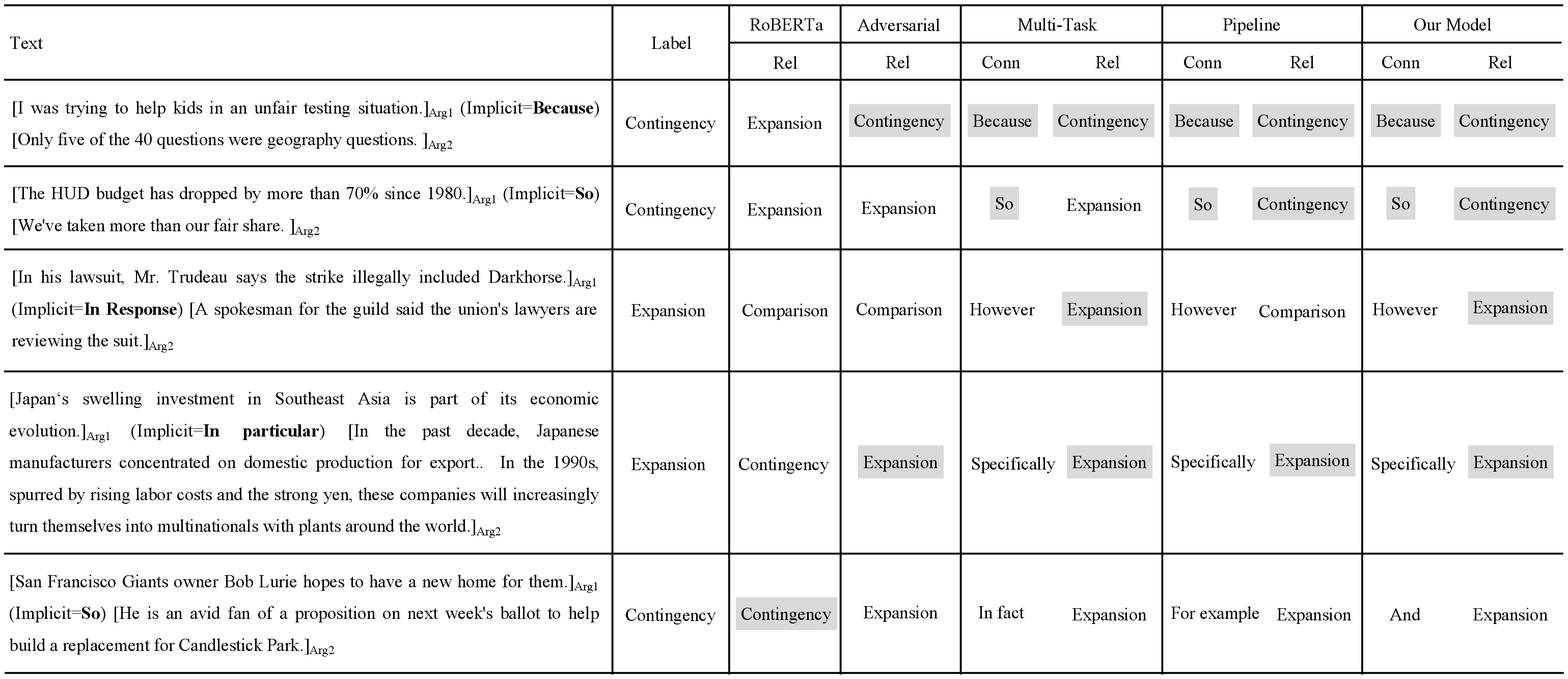}
\setlength{\abovecaptionskip}{-5pt} 
\setlength{\belowcaptionskip}{-5pt}
\caption{Examples on 11-way prediction of PDTB 2.0 from different models. RoBERTa and Adversarial can only predict relations, while Multi-Task, Pipeline, and Our Model make predictions on both connectives and relations. Therefore, we show both Connective (Conn) and relation (Rel) prediction results of the latter three models. "Text" denotes input arguments and the annotated implicit connective, and "Label" means ground truth implicit relations. Correct predictions are marked in gray background.}
\label{figure:case_study}
\end{figure*}

Table \ref{table:gt_conn} shows the Level1 classification results on PDTB 2.0 (Ji split) when manually-annotated connectives are fed to connective-enhanced models. Note that for models that do not use generated connectives, we insert the true connectives into their input in this setup. We also show the F1 results in Figure \ref{figure:pdtb2_l1_f1_inc}.

Table \ref{table:remove} shows the Level1 classification results on PDTB 2.0 when the generated connectives are removed from the inputs of relation classifiers in Pipeline and Our Method. This setting is not applied to other baselines, such as Multi-Task, because they either don't generate connectives or don't input the generated connectives into the relation classifiers. We also show the F1 results in Figure \ref{figure:pdtb2_l1_f1_rm}.

We investigate relation classifiers' performance of Multi-Task, Pipeline, and Our Model when connectives are correctly and incorrectly generated (or predicted). Other baselines, such as Adversarial, are not included in this analysis because they don't predict or generate connectives. We mentioned in Section \ref{sec:detail_ana} that Multi-task, Pipeline, and Our Model's prediction on connective are different. Specifically, their predictions do not overlap and show different performances, with a mean accuracy of 31.30\%, 33.21\%, and 32.83\% for Multi-Task, Pipeline, and Our Model, on PDTB 2.0, respectively. 

Here, we show both good and bad cases of all models from correct and incorrect connective prediction groups in Figure \ref{figure:case_study}. For comparison, we also show results from the RoBERTa and Adversarial baselines. In the first example, connective enhanced models, including Adversarial, Multi-Task, Pipeline, and Our Model, make the correct prediction on implicit relation with the help of connective information, while the RoBERTa baseline gives the wrong prediction. In the second example, Multi-Task, Pipeline, and Our Model all make the correct prediction on connectives. However, only the latter two correctly predict the implicit relations. We speculate this is because treating connectives as training objectives can not make full use of connectives. In the third example, all three models incorrectly predict the connective as "However". As a result, Pipeline incorrectly predicts the relation as "Comparison" due to the connective "However". Compared to it, both Multi-Task and Our Model correctly predict the relation "Expansion", showing better robustness. In the fourth example, all three models predict the connective as "Specifically", which is wrong but semantically similar to the manually-annotated connective "In particular". Consequently, those models all correctly predict the relation as "Expansion". In the final example, Multi-Task, Pipeline, and Our Model wrongly predict the connective as "In fact", "For example", and "And", respectively. And all three models are misled by the incorrect connectives, predicting the relation as "Expansion".

\end{document}